\documentclass[runningheads]{llncs}

\usepackage{eccv}

\usepackage{eccvabbrv}

\usepackage{graphicx}
\usepackage{booktabs}

\usepackage[accsupp]{axessibility}  

\usepackage[pagebackref,breaklinks,colorlinks]{hyperref}

\usepackage{orcidlink}

\begin{document}


\title{DualBEV: Unifying Dual View Transformation with Probabilistic Correspondences} 

\titlerunning{DualBEV}

\author{Peidong Li,
Wancheng Shen,
Qihao Huang,
Dixiao Cui}
\authorrunning{P.~Li et al.}

\institute{Zhijia Technology\\
\email{\{lipeidong, shenwancheng, huangqihao, cuidixiao\}@smartxtruck.com}}
\maketitle

\begin{abstract}
  Camera-based Bird's-Eye-View (BEV) perception often struggles between adopting 3D-to-2D or 2D-to-3D view transformation (VT). The 3D-to-2D VT typically employs resource-intensive Transformer to establish robust correspondences between 3D and 2D features, while the 2D-to-3D VT utilizes the Lift-Splat-Shoot (LSS) pipeline for real-time application, potentially missing distant information. To address these limitations, we propose DualBEV, a unified framework that utilizes a shared feature transformation incorporating three probabilistic measurements for both strategies. By considering dual-view correspondences in one stage, DualBEV effectively bridges the gap between these strategies, harnessing their individual strengths. Our method achieves state-of-the-art performance without Transformer, delivering comparable efficiency to the LSS approach, with 55.2\% mAP and 63.4\% NDS on the nuScenes test set. Code is available at \url{https://github.com/PeidongLi/DualBEV}. 
\end{abstract}

\begin{figure}[htb]
\centering
\includegraphics[width=1.0\linewidth]{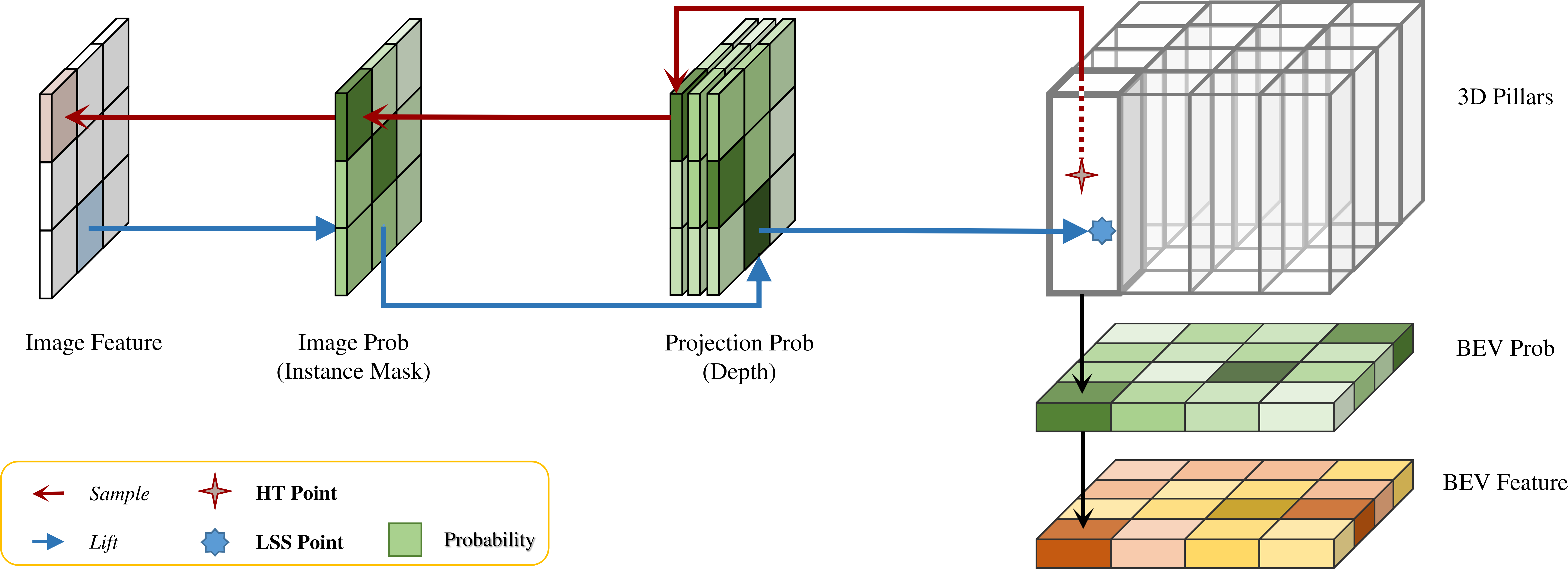}
\caption{\textbf{Unified Feature Transformation:} Our approach considers correspondences between BEV and image space utilizing image probability, projection probability and BEV probability. In the 3D-to-2D strategy, HeightTrans (HT) projects pre-defined 3D points to sample features, while in the 2D-to-3D strategy, LSS lifts image features to 3D space, both through image probability and projection probability from different directions. Finally, BEV probability is applied to enhance the representation of features.}
\label{fig:FT}
\end{figure}


\begin{figure*}[t]
    \centering
    \begin{subfigure}[b]{0.28\linewidth}
        \centering
        \includegraphics[width=\linewidth]{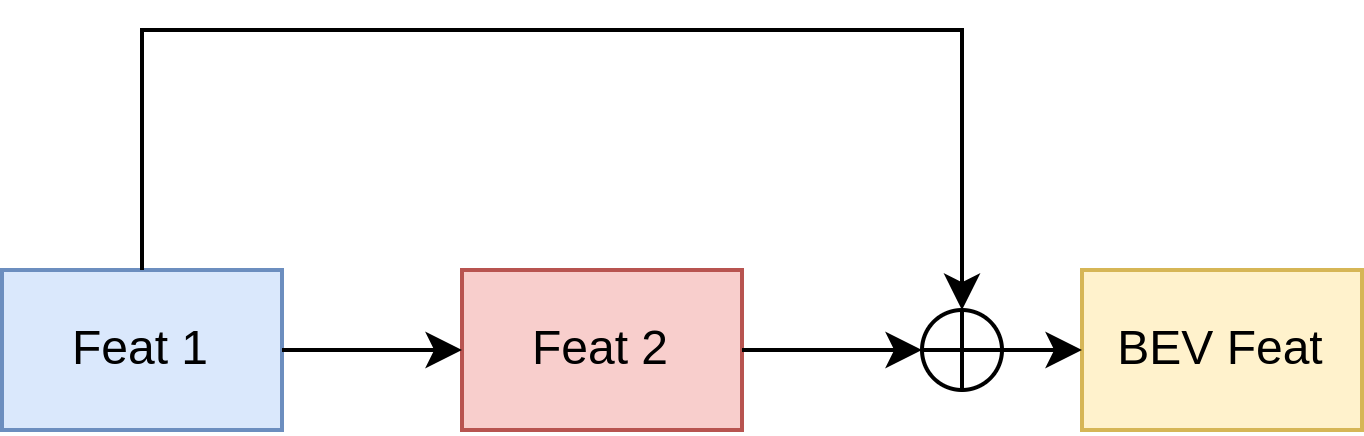}
        \caption{HeightFormer}
    \end{subfigure} \quad
    \begin{subfigure}[b]{0.28\linewidth}
        \centering
        \includegraphics[width=\linewidth]{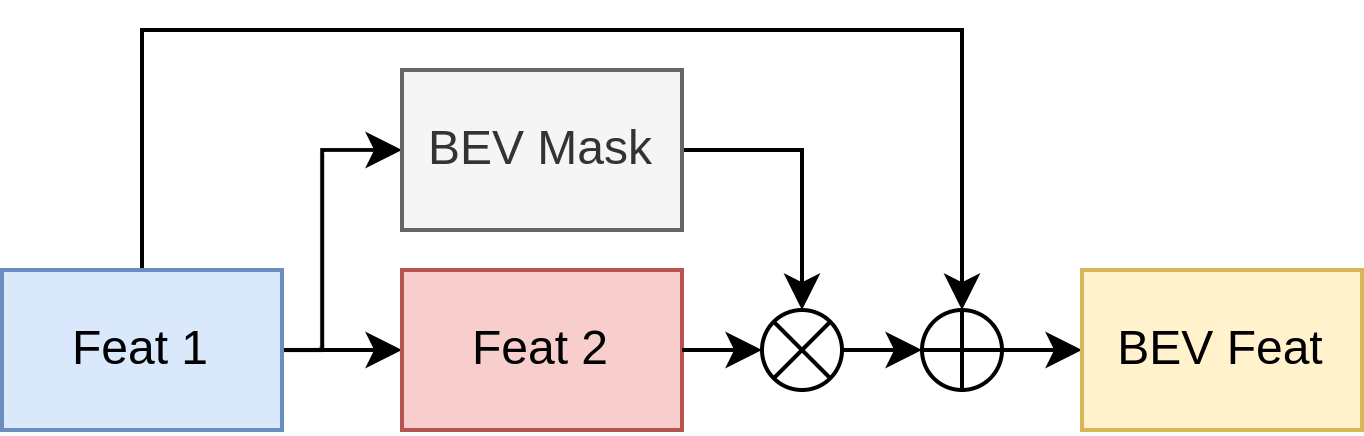}
        \caption{FB-BEV}
    \end{subfigure} \quad
    \begin{subfigure}[b]{0.28\linewidth}
        \centering
        \includegraphics[width=\linewidth]{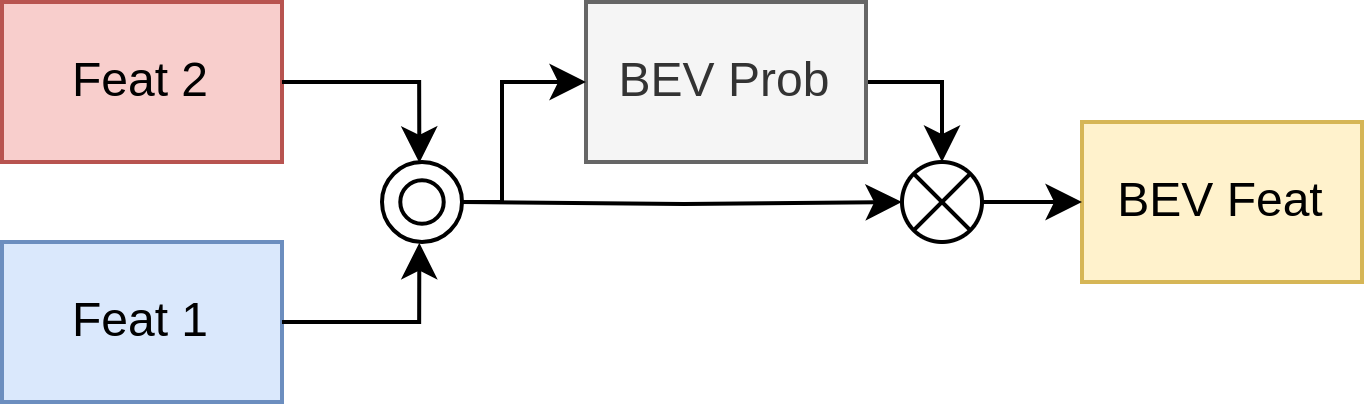}
        \caption{DualBEV}
    \end{subfigure}
    \caption{Comparison of Fusion Strategy. $\oplus$ means sum function. $\otimes$ denotes multiplication. $\circledcirc$ denotes channel-attention-based fusion.}
    \label{fig:fusion_vt}
\end{figure*}

\section{Introduction}
\label{sec:intro}

Effective BEV object detection in autonomous driving relies on precise feature transformation from the perspective to the BEV space, facilitated by the VT module. Current methods predominantly employ either a 2D-to-3D or 3D-to-2D strategy. In 2D-to-3D methods \cite{bevdet,bevdepth,bevstereo,tigbev,bevsan,sabev,matrixvt}, dense 2D features are elevated by predicting depth probabilities, but the inherent uncertainty in depth prediction can introduce inaccuracies, particularly in distant regions. Conversely, 3D-to-2D methods \cite{DETR3D,PETR,BEVFormer,polar,CAPE,heightformer,Sparse4D,liu2023sparsebev} often use BEV queries to sample 2D features, leveraging Transformer \cite{transformer} to learn attention weights for each 3D-2D correspondence, while introducing computational and deployment complexities.
HeightFormer \cite{heightformer} and FB-BEV \cite{fbbev} have explored integrating both VT. Typically, these methods employ a two-stage strategy due to different feature transformation of dual VT. Utilizing LSS features to initialize the Transformer-based VT, this strategy is constrained by the performance of the initial feature, hindering seamless fusion between the dual VT. Additionally, these methods still confront challenges in achieving real-time deployment in autonomous driving.

In this paper, we argue that both VT inherently establish correspondences between 3D and 2D features from different views, while LSS and Transformer serving as distinct methods for evaluating these correspondences. To unify dual VT, we propose a unified feature transformation (illustrated in \cref{fig:FT}) applicable to both 2D-to-3D and 3D-to-2D VT, evaluating the correspondences by three probabilistic measurements: 1) \textbf{BEV Probability}, aimed at mitigating the impact of blank BEV grids in feature construction; 2) \textbf{Projection Probability}, which distinguishes multiple correspondences, accounting for different 3D points projecting into the same 2D position; 3) \textbf{Image Probability}, aiding in excluding background features during feature transformation.

Applying this unified feature transformation, we shed light on an often overlooked approach: leveraging Convolutional Neural Network (CNN) for 3D-to-2D VT, with the introduction of \textbf{HeightTrans}. Beyond its impressive performance, we demonstrate its potential for acceleration through pre-computation, rendering it suitable for real-time autonomous driving applications. Meanwhile, we enhance the traditional LSS pipeline by integrating this feature transformation, denoted as \textbf{Prob-LSS}, showcasing its universality for current detectors. 

Combining HeightTrans and Prob-LSS, our research introduces DualBEV (see \cref{table:vtform}), an innovative approach that incorporates and considers correspondences from both BEV and perspective view in the one-stage manner, eliminating the reliance on initial features. Furthermore, we propose a robust BEV feature fusion module termed Dual Feature Fusion (DFF) module. This module enhances the integration of dual BEV features by leveraging channel attention module, while spatial attention module further helps to refine BEV probability prediction. DualBEV operates on the principle of "broad input, stringent output", leveraging precise dual-view probabilistic correspondences to comprehend and represent the probability distribution of the scene.

\begin{table}[tb]
  \caption{Comparison of different VT formulations for obtaining the BEV feature $F$ at location $(x,y)$. $p$ means point in the BEV grid, with $(u,v)$ representing the corresponding image position. $I$ denotes the image features, $D$ represents the depth maps, and $M$ signifies the instance mask of the images. $Q$ denotes BEV queries, where $B$ represents the binary BEV mask, and $P$ represents the predicted BEV probability. The notation $\mathcal{F}_{da}$ refers to the deformable attention function, while $\mathcal{G}_{2d}$ denotes the bilinear grid sampler. $f$ denotes fusion function, and $w$ represents depth consistency\cite{fbbev}. $N_z$ means number of pre-defined points in BEV grid, $N_c$ means number of corresponding points in image and $N_d$ is the number of image frustum points pooling into the grid.
  }
  \label{table:vtform}
  \centering
  \begin{tabular}{@{}l|c|l@{}}
    \toprule
    \bf{Method}  & \bf{Strategy} & \bf{VT Formulation of F(x,y)} \\
    \midrule
     BEVDet\cite{bevdet}   & 2D$\rightarrow$3D & $\sum^{N_d}_{k=1}$$D_k \cdot I_k$ \\ 
     SA-BEV\cite{sabev}    & 2D$\rightarrow$3D & $\sum^{N_d}_{k=1}$$D_k \cdot M_k \cdot I_k$ \\
     Prob-LSS(ours)    & 2D$\rightarrow$3D & $P(x,y)$ $\cdot$ $\sum^{N_d}_{k=1}$$D_k \cdot M_k \cdot I_k$ \\
     \midrule
     Simple-BEV\cite{simplebev} & 3D$\rightarrow$2D &Conv(Concat($\mathcal{G}_{2d}((u,v), I)$)) \\
     BEVFormer\cite{BEVFormer}  & 3D$\rightarrow$2D   & $\sum^{N_z}_{i=1}$$\sum^{N_c}_{j=1}$$ \mathcal{F}_{da}(Q(x,y), p_{ij}, I)$ \\
     DA-BEVFormer\cite{fbbev}  & 3D$\rightarrow$2D   & $\sum^{N_z}_{i=1}$$\sum^{N_c}_{j=1}$$ \mathcal{F}_{da}(Q(x,y), p_{ij}, I)\cdot w(D_{ij})$ \\
     HeightTrans(ours)   & 3D$\rightarrow$2D & $P(x,y)$ $\cdot$ $\sum^{N_z}_{i=1}$$\sum^{N_c}_{j=1}$$D_{ij}$ $\cdot$ $M_{ij}$ $\cdot$ $I_{ij}$ \\ 
     \midrule
     FB-BEV\cite{fbbev}   & 3D\&2D & {$\sum^{N_d}_{k=1}D_k\cdot I_k+$$B\cdot\sum^{N_z}_{i=1}\sum^{N_c}_{j=1}\mathcal{F}_{da}(Q, p_{ij}, I)\cdot w(D_{ij})$} \\ 
     DualBEV(ours)   & 3D\&2D  & $P(x,y) \cdot f(D \cdot M \cdot I)$ \\ 
  \bottomrule
  \end{tabular}
\end{table}
Our major contributions are summarized below:

\begin{itemize}
    \item [1)] We unveil the inherent similarity between 3D-to-2D and 2D-to-3D VT and identify a unified feature transformation covering dual VT. This transformation enables accurate correspondence establishment from both BEV and perspective view, significantly bridging the gap between the dual strategies.
    \item [2)] We propose a novel CNN-based 3D-to-2D VT termed HeightTrans. Leveraging probabilistic sampling and pre-computation of lookup table, HeightTrans establishes precise 3D-2D correspondences effectively and efficiently.

    \item [3)] We introduce the DFF for dual-view features fusion. This fusion strategy captures information from both close and distant regions in one-stage, culminating in the generation of comprehensive BEV features substantially.
    \item [4)] Our efficient framework DualBEV achieves state-of-the-art performance with a remarkable 55.2\% mAP and 63.4\% NDS on the nuScenes test set even without Transformer, highlighting the significance of capturing precise dual-view correspondences for view transformation.
\end{itemize}

\section{Related Work}
\subsection{3D-to-2D View Transformation}

OFT-Net\cite{oft} pioneered the integration of sampling methods into 3D-to-2D VT for monocular detection, populating voxel features by aggregating image features from corresponding projection regions. Recent BEV methods extend geometric projection with Transformer utilizing a cross-attention mechanism, broadly categorized into two streams: explicit dense BEV queries or implicit sparse object queries. The former\cite{BEVFormer,heightformer,bevformerv2} constructs a pre-defined BEV space covering a limited 3D range, with 3D-2D correspondences heavily reliant on attention mechanisms, incurring high computational costs. The latter\cite{DETR3D,Sparse4D,liu2023sparsebev} employs learnable object queries to cover all possible object proposals, a concept challenging to apply in dense tasks like lane segmentation and 3D occupancy prediction.

Contrarily, Simple-BEV\cite{simplebev} projects 3D voxels into images and samples features bilinearly, akin to OFT-Net's approach. It bypasses deformable attention\cite{deform} weights presented in Transformer-based methods, instead employing convolution to reduce concatenated channels in the height dimension. However, this simplistic sampling method without weights still lags behind 2D-to-3D CNN-based methods in latency. In this paper, we propose HeightTrans to evaluate correspondences and sum features directly in the BEV grid through a lookup table, improving the speed of this strategy significantly.

\subsection{2D-to-3D View Transformation}
A prevalent approach for 2D-to-3D VT involves lifting multi-view 2D camera features into 3D via pixel-wise discrete depth estimation, followed by BEV feature extraction through pillar sum-pooling in 3D space. This paradigmatic approach was first introduced by LSS\cite{lss} and has been followed by many subsequent works\cite{CaDDN,bevdet,bevdepth,bevstereo}. BEVDepth\cite{bevdepth} and BEVStereo\cite{bevstereo} highlight the critical role of accurate depth estimation, with explicit depth supervision enhancing performance. However, the efficiency of the subsequent $Splat$ stage remains a notable challenge, addressed by innovations such as BEV Pooling\cite{bevdet,bevdepth,bevpoolv2}. BEV-SAN\cite{bevsan} proposed a Slice Attention Module to focus on the different height slices where the different categories are located. In the concurrent work, SA-BEV\cite{sabev} suggested using SA-BEVPool to ignore points that are part of the background during BEV Pooling. Our work further extend the idea to the BEV space to ignore invalid features due to uncertainty in depth estimation.

\begin{figure}
\centering
\resizebox{\textwidth}{!}{\includegraphics{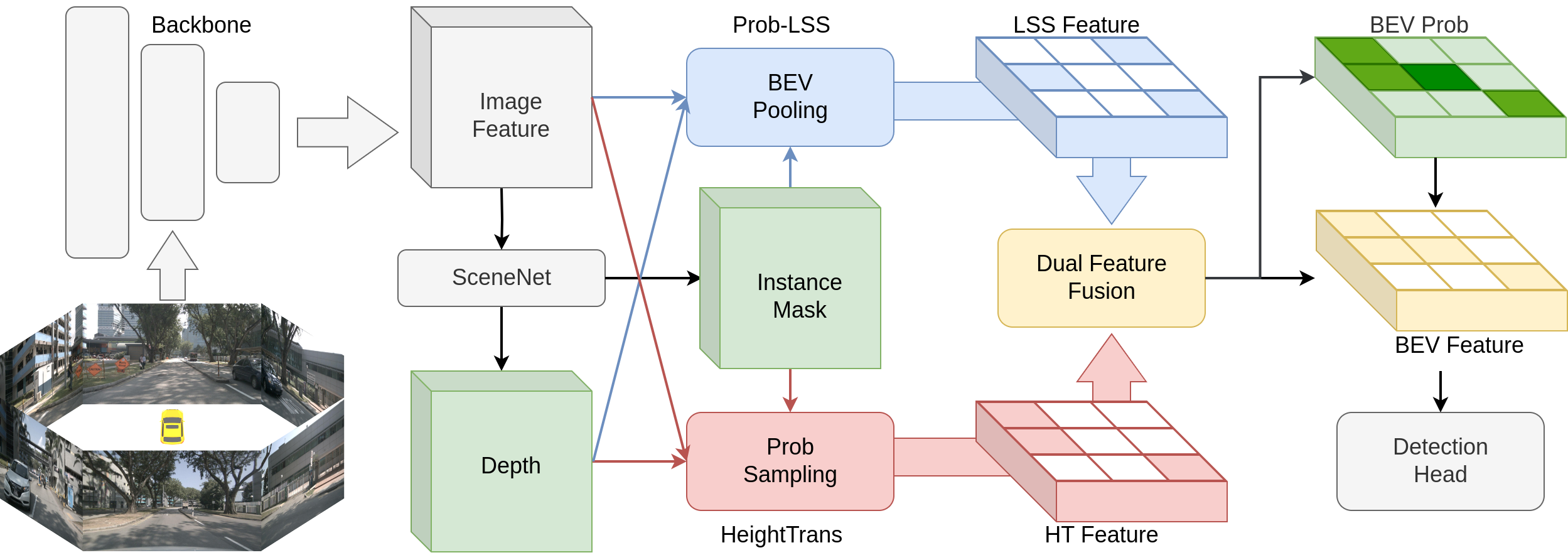}}
\caption{\textbf{Overview of DualBEV:} Initially, we employ SceneNet to predict the depth $D$ (Projection probability) and instance mask $M$ (Image probability) of input images. Subsequently, the Prob-LSS stream follows the BEVPoolv2\cite{bevpoolv2} to generate LSS feature. Concurrently, the HeightTrans stream utilizes the Prob-Sampling to project pre-defined 3D points onto the 2D space, retrieving corresponding image features. Throughout this process, all features are accompanied by probabilities derived from the depth map and instance mask. Finally, we fuse two streams and predict the BEV probability $P$ by leveraging the DFF module, resulting in the final BEV feature $F$.}
\label{fig:overview}
\end{figure}

\subsection{Fusion View Transformation}

Recent methods, such as HeightFormer\cite{heightformer} and FB-BEV\cite{fbbev}, attempt to fuse both approaches. HeightFormer introduces a height predictor on the initial BEV feature, using deformable attention for feature sampling to refine the initial feature. HeightFormer demonstrates the fusion ability by generating second-stage feature based on the first-stage feature and combining them. FB-BEV extends this concept further by introducing FRPN\cite{fbbev} to the first-stage BEV feature, selecting valid positions for Depth-Aware BEVFormer (DA-BEVFormer) to generate the second-stage feature. FB-BEV notes the differences between two paradigms, enabling a two-stage VT that leverages both 2D-to-3D and 3D-to-2D strategies. Our framework, DualBEV, further unveils the inherent sameness of dual VT, capturing information from each stream efficiently. Additionally, our method offers a more lightweight and deployment-friendly alternative without Transformer. Illustrated in \cref{fig:fusion_vt}, our approach fuses features sharing same transformation from different views in a one-stage manner, eliminating the reliance on initial BEV features and benefiting from more input information.
\section{Method}
\label{sec:Method}
As illustrated in \cref{fig:overview}, the DualBEV pipeline commences with the extraction of image feature $I\in \mathbb{R}^{N\times C_I\times H_I\times W_I}$ from $N$ cameras by the image backbone, where $H_I\times W_I$ is the shape of image feature. Subsequently, a SceneNet is employed to generate both the instance mask $M\in \mathbb{R}^{N\times C_M\times H_I\times W_I}$ and depth map $D \in \mathbb{R}^{N\times C_D\times H_I\times W_I}$. The structure of SceneNet mirrors that of DepthNet\cite{bevdepth} with just increasing output channels. Binary Cross Entropy (BCE) loss is applied for instance supervision same as depth supervision, following SA-BEV\cite{sabev}. 

The HeightTrans module employs probabilistic sampling to acquire image features. Simultaneously, the Prob-LSS stream follows the approach outlined in BEVPoolv2\cite{bevpoolv2} by lifting image with instance segmentation via depth prediction.
The features from these two streams are then fed into the DFF module for fusion and BEV probability prediction. Finally, the BEV probability $P\in \mathbb{R}^{1\times H_F\times W_F}$ is applied to the fused feature to obtain the final BEV feature $F\in \mathbb{R}^{C_F\times H_F\times W_F}$ for downstream tasks, where $H_F\times W_F$ is the shape of the BEV feature.

\subsection{HeightTrans}
\label{HT}
The fundamental principle of 3D-to-2D VT revolves around selecting 3D positions for projection into the image space and assessing these 3D-2D correspondences. In methods focusing on explicit BEV feature generation,  a pre-defined BEV map facilitates the derivation of a 2D position once the height of the BEV grid is determined. Building upon this concept, our approach begins by sampling sets of 3D points within a pre-defined BEV map. These sampled correspondences are then considered and filtered carefully to generate the BEV feature by summing within each BEV grid.

\subsubsection{BEV Height}
While existing approaches often rely on sparse uniform sampling in the height range of the BEV grid to initialize 3D points, it's important to recognize that different heights encode distinct information in the 3D space. Inspired by BEV-SAN\cite{bevsan}, HeightTrans introduces a multi-resolution sampling strategy covering the entire height range [-5m, 3m], with a resolution of 0.5m within the ROI (Region of Interest) of [-2m, 2m], and 1.0m outside this range. This sampling strategy enhances focus on small objects that might be easily missed with a coarser resolution. Unlike deformable attention methods where a set of offsets is typically predicted around the projected 2D position, our method, with its increased number of sampling points in 3D space, eliminates the need for predicting offsets in the image space, allowing for pre-computation.

\subsubsection{Prob-Sampling}
With our pre-defined set of 3D sampling points $p_{3d}\in \mathbb{R}^{3}$, the subsequent task involves acquiring features for each position and weighing the various correspondences.  Given a 3D point $p_{3d}=(x, y, z)$ in the 3D space, the camera's extrinsic matrix $T$ and intrinsic matrix $K$, the projection yields a corresponding 2D point $p_{2d}=d\cdot(u, v, 1)$ in the image space, where $d$ signifies the depth of the point.
\begin{align}
  p_{2d}=K \cdot T \cdot p_{3d}
\end{align}
A straightforward method to obtain the 3D features $F_{ht}$ is by using a bilinear grid sampler $\mathcal{G}_{2d}$ to sample the image feature $I$ at the projected position $p_{2d}$:
\begin{align}
  F_{ht}(p_{3d})=\mathcal{G}_{2d}(I, p_{2d})
\end{align}
However, the projected position may land on a background pixel, which is not only useless but could also be misleading for detection. Instead, we utilize the instance mask $M$ derived from SceneNet to represent the image probability $P_{img}$, which we apply to the image feature to mitigate this concern.
\begin{align}
  F_{ht}(p_{3d})=\mathcal{G}_{2d}(M \cdot I, p_{2d})
\end{align}
To distinguish multiple 3D points hitting the same 2D position, we further use a projection probability $P_{proj}$ to evaluate these multiple correspondences. $P_{proj}$ is obtained by a trilinear grid sampler $\mathcal{G}_{3d}$ on the depth map $D$, treating the depth channel as the third dimension.
\begin{align}
  F_{ht}(p_{3d}) = \mathcal{G}_{3d}(D, p_{2d}) \cdot \mathcal{G}_{2d}(M \cdot I, p_{2d})
\end{align}
Finally, to address the issue of blank BEV grids that offer no useful information for detection, we introduce a BEV probability $P_{bev}$ to represent the occupied probability of the BEV grid, where $(x, y)$ is the location in BEV space.
\begin{align}
  F_{ht}(p_{3d})= P(x, y) \cdot \mathcal{G}_{3d}(D, p_{2d})  \cdot \mathcal{G}_{2d}(M \cdot I, p_{2d}) \label{con:point correspondence}
\end{align}

\subsubsection{Acceleration}
BEVPoolv2\cite{bevpoolv2} utilizes pre-computation for the calculation of 3D points index in BEV space from the defined frustum, where both image feature index and depth map index remain fixed during inference. Similarly, we can accelerate our VT by building a lookup table after replacing the grid sampler with the round function. The BEV feature in \cref{con:point correspondence} can be simplified as:
\begin{align}
  F_{ht}(x,y,z)=P(x, y) \cdot D(u, v, d) \cdot M(u, v) \cdot I(u, v) \label{eq:Fht}
\end{align}

Now we can sum the features of $N_z$ pre-defined points in each BEV grid for $N_c$ corresponding 2D positions to obtain the final HeightTrans feature as:
\begin{align}
  F_{ht}(x,y)=P(x, y) \cdot \sum^{N_z}_{i=1}\sum^{N_c}_{j=1} D(u_{ij}, v_{ij}, d_{ij}) \cdot M(u_{ij}, v_{ij}) \cdot I(u_{ij}, v_{ij}) \label{eq:Fht}
\end{align}
As shown in \cref{table:vtform}, this representation bears similarity to BEV Pooling. Therefore, we can establish the lookup table by projecting pre-defined 3D points into the image space and subsequently calculating the index in the feature map and depth map. Unlike BEV Pooling, the index in BEV map is constant in our method. Then we can use the same CUDA operator as in BEVPoolv2\cite{bevpoolv2} with the lookup table to speed up the calculation of BEV features during inference. 

\subsection{Prob-LSS}
The traditional LSS pipeline begins by predicting the depth probability of each pixel to facilitate its elevation into frustum points, which is then projected into the BEV space through BEV Pooling. SA-BEV\cite{sabev} introduced Semantic-Aware BEV Pooling to circumvent the need for lifting irrelevant pixels in image space. However, the inherent uncertainty in depth estimation may still result in extraneous information being present in the BEV space.

To address this issue, we further integrate BEV probability into the LSS pipeline, referred to as Prob-LSS. The construction of LSS features at location $(x,y)$ is similar to \cref{eq:Fht} and can be expressed as follows:
\begin{align}
  F_{lss}(x,y)=P(x, y) \cdot \sum^{N_d}_{k=1} D(u_k, v_k, d_k) \cdot M(u_k, v_k) \cdot I(u_k, v_k) 
  \label{eq:Flss}
\end{align}
where $N_d$ represents the number of projected frustum points in the BEV grid at location $(x, y)$. It's noteworthy that for each BEV grid, HeightTrans provides a constant number with $N_z \times N_c$ transformed image features, whereas LSS offers a dynamic number with $N_d$ features depending on the lifted frustum. This dynamic nature of LSS acts as a complementary aspect to the HeightTrans.
\begin{figure}[tb]
    \centering
    \begin{subfigure}{0.3\linewidth}
      \resizebox{\textwidth}{!}{\includegraphics{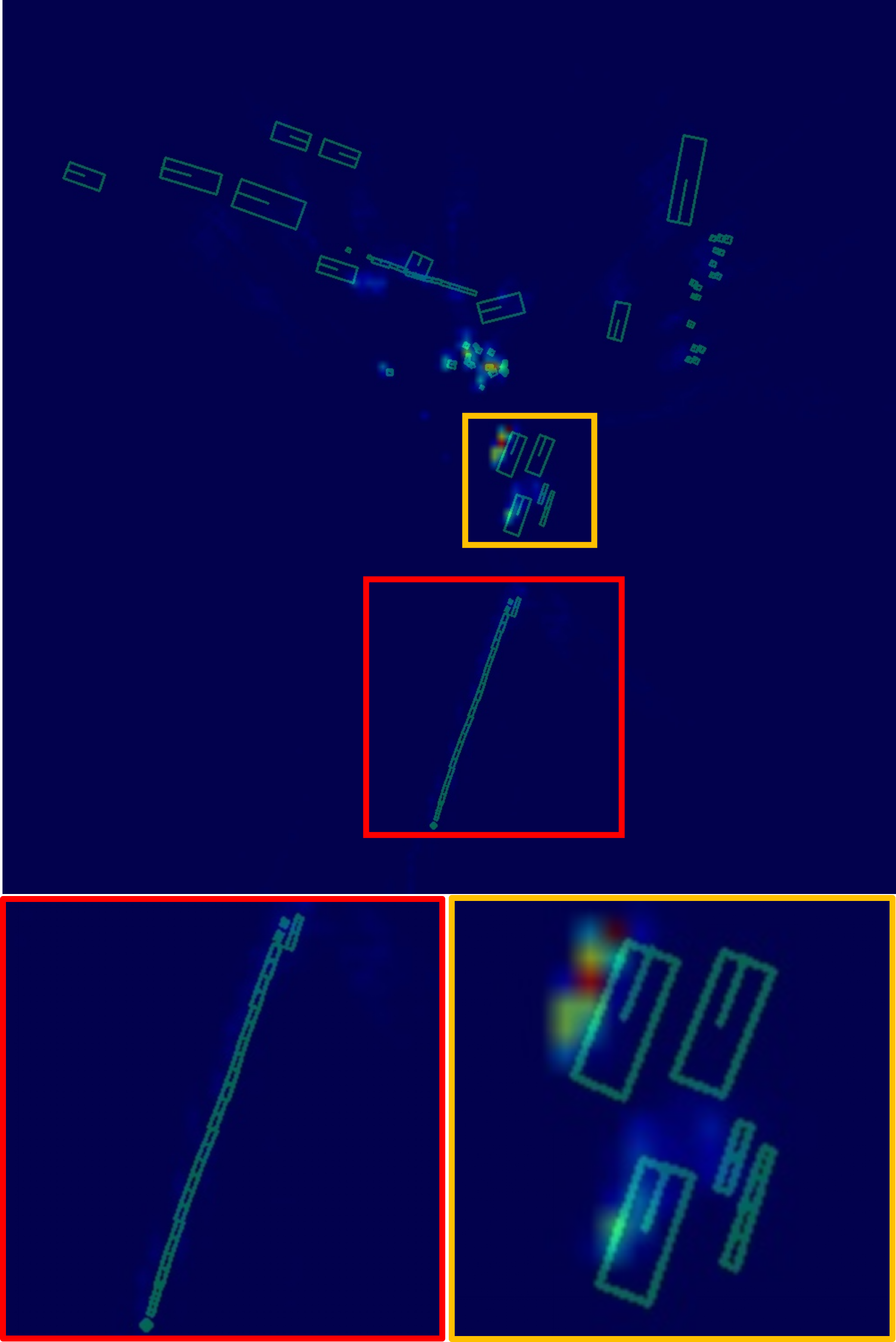}}
      \caption{Prob-LSS Feature}
      \label{fig:lss_vis}
    \end{subfigure}
    \begin{subfigure}{0.3\linewidth}
      \resizebox{\textwidth}{!}{\includegraphics{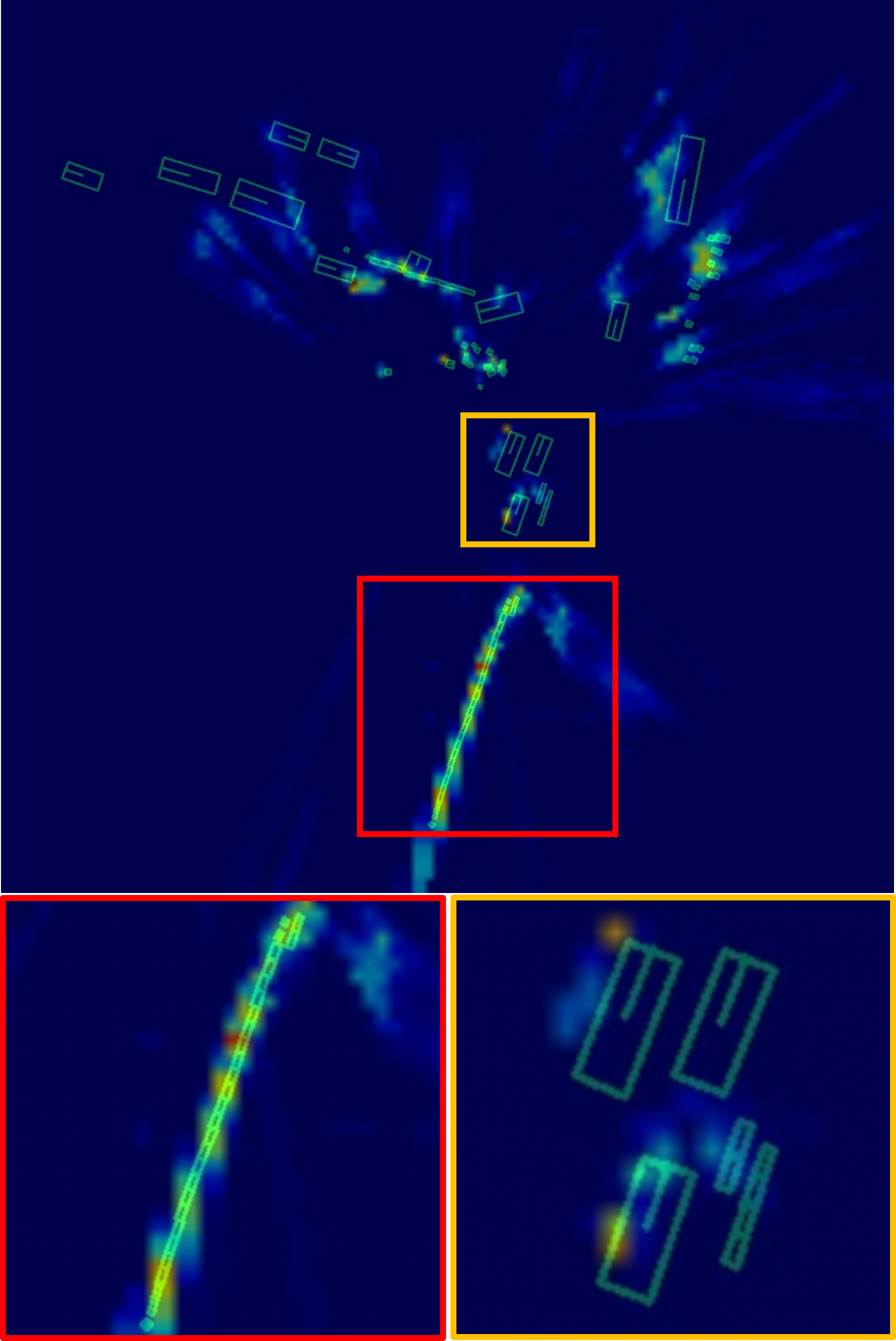}}
      \caption{HeightTrans Feature}
      \label{fig:ht_vis}
    \end{subfigure}
    \begin{subfigure}{0.3\linewidth}
        \resizebox{\textwidth}{!}{\includegraphics{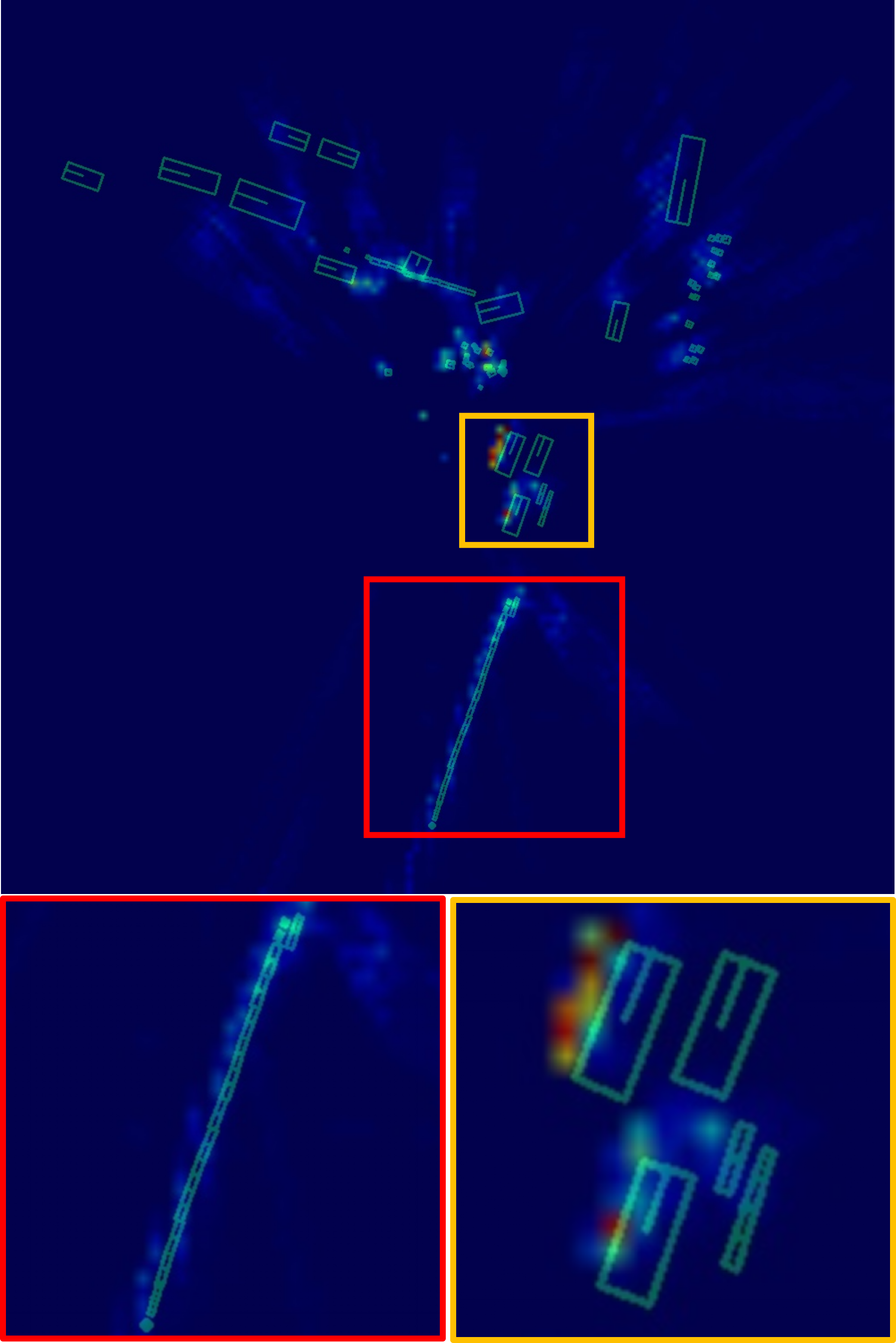}}
        \caption{Fused Feature}
        \label{fig:fus_vis}
    \end{subfigure}
    \caption{\textbf{BEV Feature Visualization with GT boxes.} Prob-LSS pays more attention to close range while HeightTrans can also capture distant information. In the red rectangle (distant range), where our unified framework compensates weak detection on barriers of Prob-LSS with HeightTrans. In the orange rectangle (close range), BEV features are enhanced from dual streams. Ego locates in the center of BEV features.}
    \label{fig:feat_vis}
\end{figure}

\subsection{Dual Feature Fusion}
\label{DFF}
\begin{figure}
\centering
\resizebox{\textwidth}{!}{\includegraphics{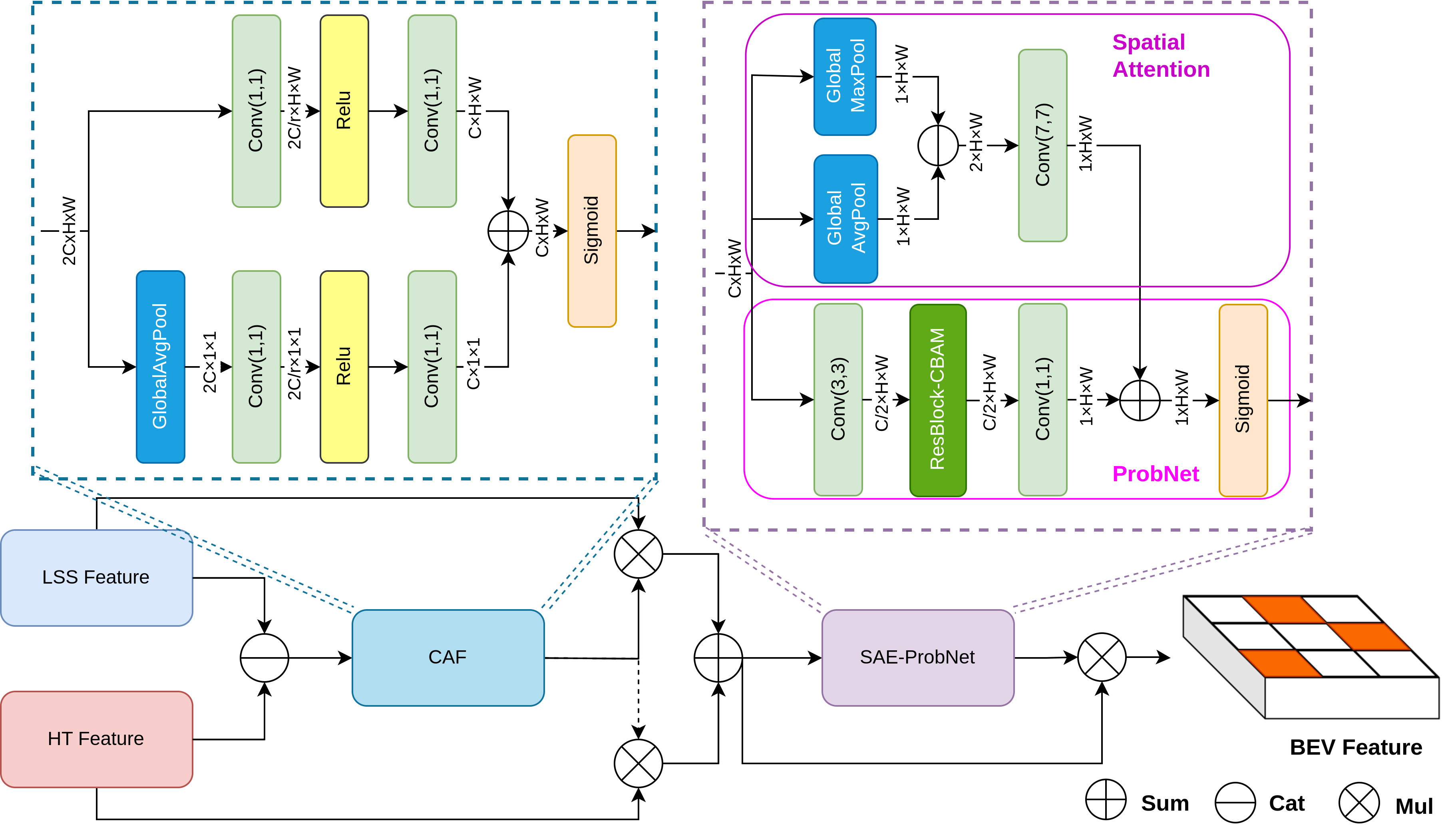}}
\caption{\textbf{Dual Feature Fusion Module}: Dual features are first concatenated and then passed into the CAF module for fusion. Subsequently, the SAE-ProbNet is utilized to obtain the BEV probability for the final BEV feature.}
\label{fig:dff}
\end{figure}
Unifying \cref{eq:Fht,eq:Flss}, we observe that BEV probability can be obtained after the fusion of dual features. Inspired by the CBAM\cite{cbam} and AFF\cite{aff} framework, we propose a Dual Feature Fusion (DFF) module to integrate these features and predict the BEV probability effectively. The DFF module comprises a fusion module $f$, which utilizes channel attention to predict weights for dual feature summation, and a Spatial Attention Enhanced ProbNet (SAE-ProbNet) to predict the BEV probability $P$, formulated as:
\begin{align}
  F(x,y)=P(x, y) \cdot f(F_{lss}(x,y), F_{ht}(x,y))
\end{align}
\subsubsection{Fusion Module}
As shown in \cref{fig:feat_vis}, $F_{lss}$ focuses more on close objects occupying most of the image pixels, while $F_{ht}$ tends to contain information of distant objects. This observation suggests two-stage methods\cite{heightformer,fbbev} initialized by LSS features, may not guide second-stage VT to extract features at distant regions.

To tackle these challenges and obtain a robust BEV representation, our fusion module $f$ concatenates the dual features and passes them into a channel-attention-based fusion (CAF) module to predict the affinity for feature selection. The fusion feature $F_{channel}$ is obtained using the following equation:
\begin{align}
  F_{channel}&=f(F_{lss}, F_{ht}) \\
  &=C(F_{lss}\ominus F_{ht}) \cdot F_{lss} + (1-C(F_{lss}\ominus F_{ht})) \cdot F_{ht}
\end{align}
where $\ominus$ denotes concatenation, and \textit{C} denotes the CAF module, which is modified from the MS-CAM\cite{aff} as illustrated in \cref{fig:dff}. This fusion phase aims to select features from two streams softly with a learning weight, enhancing the representation in both close and distant regions.

\subsubsection{BEV Probability Prediction}
We utilize the BEV probability $P$ predicted by the SAE-ProbNet to aggregate $F_{channel}$, thereby mitigating the impact of blank BEV grids. We employ ProbNet to extract local information, serving as the local stream $P_l$. Further enhancement of the prediction is achieved by incorporating a spatial attention module to capture global information, serving as the global stream $P_g$. The entire module can be formulated as:


\begin{align}
  F_{spatial}&=P\cdot F_{channel} \\
  &= \sigma(P_l(F_{channel})+P_g(F_{channel})) \cdot F_{channel} 
\end{align}

As depicted in \cref{fig:dff}, the local stream, ProbNet, employs a 3$\times$3 convolution kernel to reduce the channel dimensions. Subsequently, it undergoes processing through a ResBlock-CBAM\cite{cbam} and a 1$\times$1 convolution operation to acquire local attention. Supervised by a BEV mask, ProbNet utilizes BCE loss and Dice loss\cite{vnet}, with BEV Centerness\cite{m2bev} also incorporated into the loss to encourage the network to focus more on distant objects. On the other hand, the global stream utilizes a 7$\times$7 convolution kernel to compute the average and maximum values of the input feature, thus enlarging the perception field in BEV space. These streams are then combined before passing through the sigmoid function $\sigma$. This design aims to equip our module with the capability to capture both local and global attention for enhanced BEV probability prediction.

\begin{table}[tb]
  \caption{Comparison on the nuScenes val set. ResNet-50 is used as image backbone and resolution is set to 704$\times$256. $\dag$: We apply BEVDepth\cite{bevdepth} with BEVPoolv2\cite{bevpoolv2} as baseline method. $\star$: For fair comparison with LSS method, we disable SceneNet and instance supervision in single-frame. $\ddag$: Features' operation is performed at 1/16 resolution without BEV-Paste\cite{sabev} for fair comparison.
  }
  \label{tab:r50}
  \centering
  \scalebox{0.9}{
  \begin{tabular}{@{}l|c|cc|ccccc@{}}
    \toprule
    Method  & Frames & mAP$\uparrow$ & NDS$\uparrow$  & mATE$\downarrow$   & mASE$\downarrow$   & mAOE$\downarrow$   & mAVE$\downarrow$   & mAAE$\downarrow$\\
    \midrule
    BEVDet\cite{bevdet}     & 1  & 29.8  & 37.9 & 0.725 & 0.279 & 0.589 & 0.860 & 0.245 \\
    BEVDepth\cite{bevdepth}     & 1  & 33.7  & 41.4 & 0.646 & 0.271 & 0.574 & 0.838 & 0.220 \\
    BEVDepth\cite{bevpoolv2} $\dag$     & 1  & 34.2  & 40.7 & 0.645 & 0.273 & 0.599 & 0.890 & 0.240 \\
    \textbf{DualBEV}$\star$       & 1  & \textbf{35.2}  & \textbf{42.5} & 0.640 & 0.271 & 0.542 & 0.838 & 0.216 \\
    \hline
    BEVDet4D\cite{bevdet4d}      & 2  & 32.2  & 45.7 & 0.511 & 0.241 & 0.386 & 0.301 & 0.121 \\
    BEVFormer\cite{BEVFormer}      & 2  & 33.0 & 45.9 & 0.686 & 0.272 & 0.482 & 0.417 & 0.201 \\
    BEVDepth\cite{bevdepth}      & 2  & 35.1  & 47.5 & 0.639 & 0.267 & 0.479 & 0.428 & 0.198 \\
    BEVDepth$\dag$\cite{bevpoolv2}       & 2  & 36.8  & 48.5 & 0.609 & 0.273 & 0.507 & 0.406 & 0.196 \\
    BEVStereo\cite{bevstereo}      & 2  & 37.2  & 50.0 & 0.598 & 0.270 & 0.438 & 0.367 & 0.190 \\
    FB-BEV\cite{fbbev}      & 2  & 37.8  & 49.8 & 0.620 & 0.273 & 0.444 & 0.374 & 0.200 \\
    SA-BEV$\ddag$\cite{sabev}     & 2  & 37.8  & 49.9 & 0.617 & 0.270 & 0.441 & 0.370 & 0.206 \\
    \textbf{DualBEV}      & 2  & \textbf{38.0}  & \textbf{50.4} & 0.612 & 0.259 & 0.403 & 0.370 & 0.207
\\ 
  \bottomrule
  \end{tabular}
  }
\end{table}

\begin{table}[t]
\caption[l]{
{Comparison with previous state-of-the-art BEV detectors on the nuScenes test set. Swin-B\cite{swin} and ConvNeXt-B\cite{convnext} don't use extra data for depth training.}}
\label{table:detector}
\centering
\scalebox{0.9}{
\begin{tabular}{l|c|cc|ccccc}
\toprule
Method  & Backbone & mAP$\uparrow$ & NDS$\uparrow$  & mATE$\downarrow$   & mASE$\downarrow$   & mAOE$\downarrow$   & mAVE$\downarrow$   & mAAE$\downarrow$ \\ \midrule[0.8pt]
DETR3D\cite{DETR3D}     & V2-99 & 41.2  & 47.9 & 0.641 & 0.255 & 0.394 & 0.845 & 0.133 \\
BEVDet4D\cite{bevdet4d}       & Swin-B  & 45.1  & 56.9 & 0.511 & 0.241 & 0.386 & 0.301 & 0.121 \\
UVTR\cite{uvtr}     & V2-99 & 47.2  & 55.1 & 0.577 & 0.253 & 0.391 & 0.508 & 0.123 \\
BEVFormer\cite{BEVFormer}       & V2-99  & 48.1  & 56.9 & 0.582 & 0.256 & 0.375 & 0.378 & 0.126 \\
BEVDepth\cite{bevdepth}       & V2-99 & 50.3  & 60.0 & 0.445 & 0.245 & 0.378 & 0.320 & 0.126 \\
PETRv2\cite{petrv2}       & V2-99 & 50.8  & 59.1 & 0.543 & 0.241 & 0.360 & 0.367 & 0.118 \\
Sparse4D\cite{Sparse4D}       & V2-99 & 51.1  & 59.5 & 0.533 & 0.263 & 0.369 & 0.317 & 0.124 \\
BEVStereo\cite{bevstereo}       & V2-99 & 52.5  & 61.0 & 0.431 & 0.246 & 0.358 & 0.357 & 0.138 \\
SA-BEV\cite{sabev}       & V2-99 & 53.3  & 62.4 & 0.430 & 0.241 & 0.338 & 0.282 & 0.139 \\
FB-BEV\cite{fbbev}       & V2-99 & 53.7  & 62.4 & 0.439 & 0.250 & 0.358 & 0.270 & 0.128 \\
SOLOFusion\cite{solofusion}       & ConvNeXt-B  & 54.0  & 61.9 & 0.453 & 0.257 & 0.376 & 0.276 & 0.148 \\
SparseBEV\cite{liu2023sparsebev}       & V2-99 & 54.3  & 62.7 & 0.502 & 0.244 & 0.324 & 0.251 & 0.126 \\
\hline
\textbf{DualBEV}       & V2-99  & \textbf{55.2}  & \textbf{63.4} & 0.414 & 0.245 & 0.377 & 0.252 & 0.129 \\ 
\bottomrule
\end{tabular}
}
\end{table}
\section{Experiments}
\subsection{Datasets and Metrics}
We conduct our experiments on the nuScenes dataset\cite{nuscenes}, a widely used benchmark for autonomous driving research. nuScenes offers comprehensive sensor data captured in urban driving scenarios, facilitating robust evaluation of object detection algorithms. NuScenes Detection Score (NDS)\cite{nuscenes} serves as the official metric to measure the quality of 3D detection. NDS integrates mean Average Precision (mAP), mean Average Translation Error (mATE), mean Average Scale Error (mASE), mean Average Orientation Error (mAOE), mean Average Velocity Error (mAVE) and mean Average Attribute Error (mAAE), providing a holistic measure of detection quality across various aspects of performance.

\subsection{Implementation Details}
We employ ResNet-50\cite{resnet} operating at an image resolution of 704$\times$256. During the 20-epoch training phase with the CBGS~\cite{cbgs} and a batch size of 64, we utilize the AdamW~\cite{adamw} optimizer with a learning rate set to 2$\times10^{-4}$. Data augmentation techniques consistent with those of BEVDet~\cite{bevdet} are applied. 
Our BEV grid size is set to 128$\times$128 without building voxel features. For the test set, we leverage VoVNet~\cite{vovnet} at an image resolution of 1600$\times$640, with the BEV grid size adjusted to 256$\times$256. The model is trained for only 8 epochs with CBGS, and 8 previous keyframes are applied for temporal module following BEVDet4D\cite{bevdet4d}. All latency tests are conducted on a single NVIDIA 3090 GPU.

\subsection{Main Results}
\subsubsection{nuScenes val set}
We adopt BEVDepth\cite{bevdepth} with BEVPoolv2\cite{bevpoolv2} as baseline method. As illustrated in \cref{tab:r50}, DualBEV outperforms the baseline method with +1.0\% mAP and +1.8\% NDS in single frame. For multi-frame, DualBEV not only surpasses the baseline with +1.2\% mAP and +1.9\% NDS, but also outperforms BEVStereo\cite{bevstereo} which uses temporal-stereo module for precise depth estimation. By leveraging HT features which capture the distant information, and efficient fusion with LSS feature, we also have a 0.5\% NDS increase compared to SA-BEV\cite{sabev} and a 0.6\% NDS increase compared to FB-BEV\cite{fbbev}.

\subsubsection{nuScenes test set}
\cref{table:detector} demonstrates that DualBEV achieves state-of-the-art performance, yielding a remarkable 55.2\% mAP and 63.4\% NDS compared to prior works in VT. Notably, our approach exhibits a significant margin of 1.9\% in mAP over SA-BEV\cite{sabev} and 1.5\% over FB-BEV\cite{fbbev}, while also surpassing them by 1.0\% in NDS. Additionally, when compared to SOLOFusion\cite{solofusion}, which focus on temporal module, DualBEV maintains superiority with a lead of 1.2\% in mAP and 1.5\% in NDS even without temporal-stereo module. Furthermore, DualBEV outperforms the previous state-of-the-art Transformer-based VT approach SparseBEV\cite{liu2023sparsebev}, by a notable margin of 0.9\% in mAP and 0.7\% in NDS, proving the pivatol role of unveiling precise correspondences in VT.
\begin{table}[tb]
    \caption[l]{{Ablation study of different components on the nuScenes val set.}}
    \label{tab:Ablation Studt}
    \centering
    \begin{subtable}{0.55\linewidth}
        \centering
        \begin{tabular}{l|cc|r}
        \toprule
        Method   & mAP$\uparrow$ & NDS$\uparrow$ & latency$\downarrow$  \\ \midrule[0.8pt]
        Baseline\cite{bevdepth}     & 34.2  & 40.7  & 10.2ms\\ 
        +ProbNet      & 34.6  & 41.1   & 11.1ms\\ 
        +HeightTrans      & 35.0  & 41.5  & 11.3ms \\ 
        +CAF      & \bf{35.3}  & 41.6   & 11.5ms\\ 
        +SAE     & 35.2  & \bf{42.5}  & 11.7ms\\ 
        \bottomrule
        \end{tabular}
        \caption{Component Ablation.}
        \label{tab:Component Ablation}
    \end{subtable}
    \quad
    \begin{subtable}{0.4\linewidth}
         \centering
        \begin{tabular}{ccc|cc}
        \toprule
        $P_{proj}$ &$P_{img}$ &$P_{bev}$  & mAP$\uparrow$ & NDS$\uparrow$ \\ \midrule[0.8pt]
        $\checkmark$ &   &   & 37.0  & 48.8  \\
        $\checkmark$ & $\checkmark$ &   & 37.7  & 49.9  \\
        $\checkmark$ &   & $\checkmark$ & 37.4  & 49.6  \\
          & $\checkmark$ & $\checkmark$ & 27.0  & 40.0 \\
        $\checkmark$ & $\checkmark$ & $\checkmark$ & \bf{38.0}  & \bf{50.4} \\
        \bottomrule 
        \end{tabular}
        \caption{Probability Ablation.}
        \label{tab:Probability Ablation}
    \end{subtable}
    \\
    \begin{subtable}{0.55\linewidth}
    \centering
        \begin{tabular}{l|cc|r}
        \toprule
        VT Operation   & mAP$\uparrow$ & NDS$\uparrow$ & latency$\downarrow$  \\ \midrule[0.8pt]
        $\mathrm{SCA}_{da}$\cite{fbbev} & 29.5  & 36.6 & 12.50ms \\ 
        Bilinear-Sampling\cite{simplebev} & 30.5  & 37.6 & 7.15ms \\ 
        
        BEV Pooling\cite{bevpoolv2} & \bf{30.7}  & 38.2  & 0.32ms \\ 
        Prob-Sampling & 30.6  & \bf{39.0} & 0.32ms \\ 
        \bottomrule
        \end{tabular}
        \caption{VT Operation Comparison.}
        \label{tab:VT Comparison}
    \end{subtable}
    \quad
    \begin{subtable}{0.4\linewidth}
        \centering
        \begin{tabular}{c|c|cc}
        \toprule
        Distribution  & Num & mAP$\uparrow$ & NDS$\uparrow$  \\ \midrule[0.8pt]
        Uniform     & 4 & 35.0  & 41.9  \\
        Uniform     & 8 & 35.3  & 42.2  \\
        Uniform    & 16 & \bf{35.3}  & 42.3  \\
        MR    & 13 & 35.2  & \bf{42.5}  \\
        \bottomrule
        \end{tabular}
        \caption{Sampling Strategy in Height.}
        \label{tab:Sampling Strategy}
    \end{subtable}
    \\
    \begin{subtable}{0.55\linewidth}
        \centering
        \begin{tabular}{l|ccc|cc}
        \toprule
        Method &$P_{proj}$ &$P_{img}$ &$P_{bev}$   & mAP$\uparrow$ & NDS$\uparrow$ \\ \midrule[0.8pt]
        BEVDet\cite{bevdet} & $\checkmark$ &  &  & 30.7  & 38.2  \\
        SA-BEV\cite{sabev} & $\checkmark$ & $\checkmark$ &   & 30.9  & 38.7 \\
        Prob-LSS & $\checkmark$ & $\checkmark$ & $\checkmark$ & \bf{31.5}  & \bf{39.1} \\
        \bottomrule
        \end{tabular}
        \caption{Prob-LSS Ablation.}
        \label{tab:Stream Ablation}
    \end{subtable}
    \quad
    \begin{subtable}{0.4\linewidth}
        \centering
        \begin{tabular}{c|c|cc}
        \toprule
        Strategy  &Process & mAP$\uparrow$ & NDS$\uparrow$   \\ \midrule[0.8pt]
        Refine   & two-stage & 35.1  & 41.7  \\
        Add      & one-stage & 35.2  & 42.1  \\
        DFF     & one-stage & \bf{35.2}  & \bf{42.5}  \\
        \bottomrule
        \end{tabular}
        \caption{Fusion Strategy.}
        \label{tab:Fusion Strategy}
    \end{subtable}
\end{table}

\subsection{Ablation Study}
\subsubsection{Impact of Each Component}
We first explore each component's impact in \cref{tab:Ablation Studt}(a). Starting with BEVDepth\cite{bevdepth} as the baseline, the initial integration of ProbNet to provide BEV probability results in a 0.4\% increase in both mAP and NDS. Combining HT features led to a 0.4\% mAP and 0.4\% NDS increase, showcasing compensation for Prob-LSS. Subsequently, we employ the CAF module to softly select features from the two streams, resulting in a further 0.3\% mAP and 0.1\% NDS increase. Finally, by incorporating the SAE module for improved BEV probability prediction with the addition of spatial attention module alongside ProbNet, our method achieves a substantial 0.9\% improvement in NDS. The overall components gain a 1.0\% mAP and 1.8\% NDS improvement. Additionally, we also test the latency of VT from image feature to BEV feature including depth estimation. The entire proposed components reveal an extra latency of merely 1.5ms, with ProbNet introducing 0.9ms of this total.

\subsubsection{Impact of Each Probability}
\cref{tab:Ablation Studt}(b) demonstrates the impact of each probabilistic measurement for DualBEV. Alongside the projection probability $P_{proj}$, the inclusion of image probability $P_{img}$ yields a notable improvement of 0.7\% mAP and 1.1\% NDS, while the introduction of BEV probability $P_{bev}$ contributes to a 0.4\% mAP and 0.8\% NDS enhancement. This observation implies that features of background pixels in the image space have a comparable detrimental effect to features of blank grids in the BEV space. An even more significant performance boost of 1.0\% mAP and 1.6\% NDS is achieved when both probabilities are applied simultaneously. However, disabling $P_{proj}$ by adopting a uniform distribution for depth prediction per pixel leads to a significant decrease in performance, indicating the critical importance of $P_{proj}$, which serves as the cornerstone of the LSS pipeline.
\subsubsection{Effect of Prob-Sampling}

In \cref{tab:Ablation Studt}(c), we compare Prob-Sampling with other methods using BEVDet\cite{bevdet} as the base detector, focusing solely on VT operation. Regarding accuracy, Prob-Sampling outperforms 1-layer depth-aware spatial cross-attention ($\mathrm{SCA}_{da}$) proposed by FB-BEV\cite{fbbev} by 2.4\% in NDS, while achieving a 1.4\% NDS enhancement compared to bilinear sampling used in Simple-BEV\cite{simplebev}. Additionally, Prob-Sampling shows a 0.8\% NDS improvement over BEV Pooling\cite{bevpoolv2}. In terms of inference latency, Prob-Sampling's adoption of pre-computation for acceleration allows it to achieve comparable speed to BEV Pooling, representing a remarkable more than 20 times speed improvement over bilinear sampling and 40 times improvement over $\mathrm{SCA}_{da}$. This highlights Prob-Sampling's competitiveness in both accuracy and latency for BEV detection. 
\subsubsection{Effect of Sampling Strategy in Height}
In HeightTrans, we apply a multi-resolution (MR) sampling strategy within the height range. We evaluate different sampling strategies as shown in \cref{tab:Ablation Studt}(d). The performance almost reaches saturation after 8 uniform points. Notably, our proposed approach achieves a 0.6\% increase in NDS compared to 4 uniform points utilized in BEVFormer\cite{BEVFormer}, and even a 0.2\% enhancement compared to 16 uniform points along the height.
\subsubsection{Effect of Prob-LSS}
Our proposed Prob-LSS extends the representation of LSS approach as illustrated in \cref{table:vtform}. We further examine the evolution of LSS methods starting from BEVDet\cite{bevdet} without any auxiliary loss in \cref{tab:Ablation Studt}(e). SA-BEV\cite{sabev} expands upon the base version by introducing instance segmentation to filter out irrelevant information in the image space. This addition results in a 0.2\% increase in mAP and a 0.5\% increase in NDS. Our method further extends the idea to BEV space to mitigate the impact of irrelevant BEV features, particularly those generated by inaccurate depth estimation. This extension leads to a notable improvement of 0.6\% in mAP and 0.4\% in NDS.
\subsubsection{Effect of Fusion Strategy}
\cref{tab:Ablation Studt}(f) compares different fusion strategy for HT feature and Prob-LSS feature. We first explore the refinement strategy employed in FB-BEV\cite{fbbev}, which utilizes Prob-LSS features to predict the BEV probability for HeightTrans. However, the refinement strategy yields even worse results than directly summing the both features for BEV probability prediction, with a 0.4\% NDS decrease. Our proposed DFF strategy achieves a notable improvement, surpassing the refinement strategy by 0.8\% in NDS.

 
\begin{figure}
\centering
\resizebox{\textwidth}{!}{\includegraphics{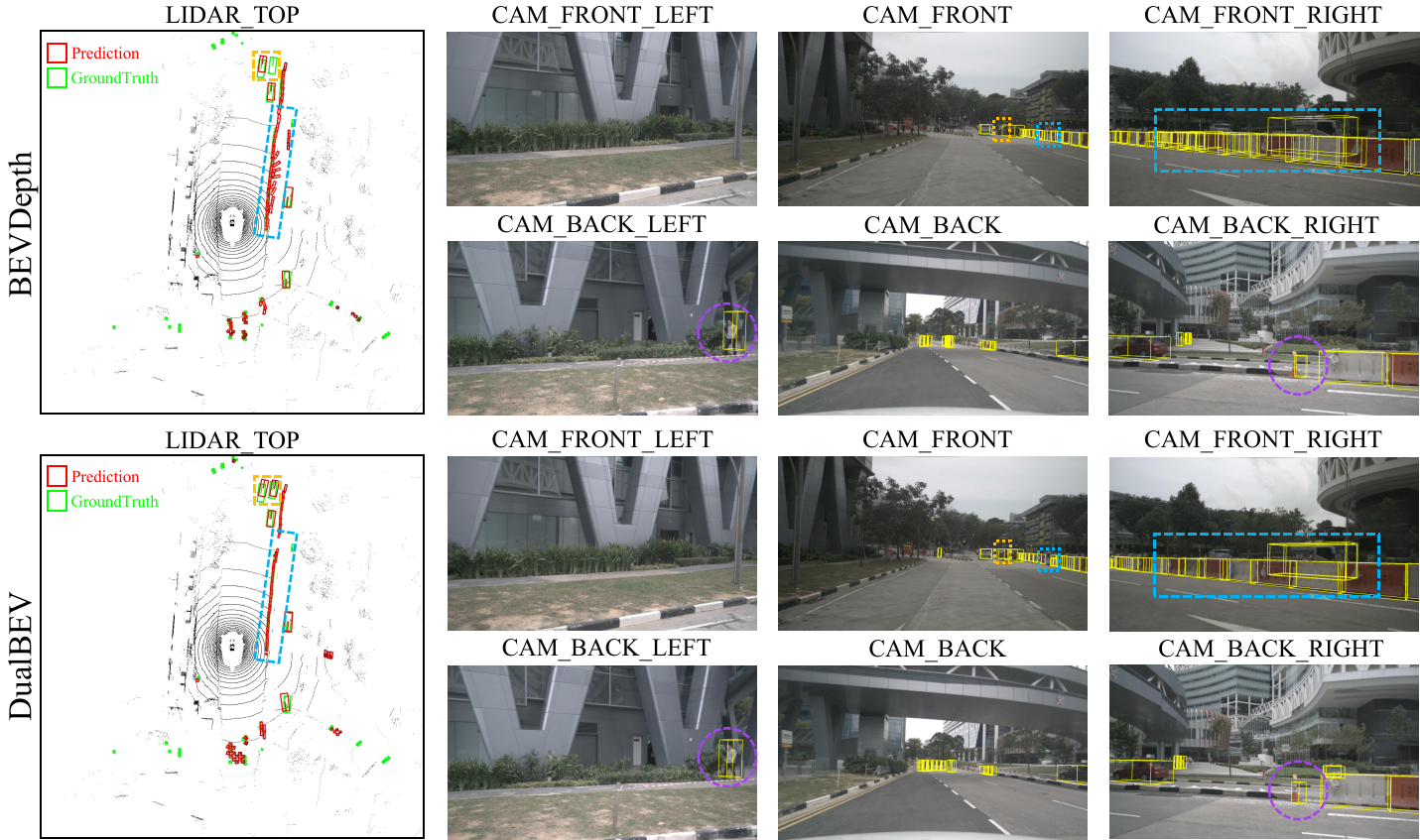}}
\caption{Visualization of detection results on image and BEV space.}
\label{fig:test}
\end{figure}

\subsection{Visualization}
In \cref{fig:test}, we present a qualitative comparison between BEVDepth\cite{bevdepth} and DualBEV. Our approach exhibits superior performance in close and medium ranges, effectively eliminating false detections and accurately captures the curve of barriers even break (blue dashed rectangles). In the distant range, our method also recalls missed objects (orange dashed rectangles), benefiting from the compensation of HeightTrans. Additionally, our method provides precise information about small objects (purple dashed circles), which is not evident from BEV.

\section{Conclusion and Limitation}
In this work, we present a novel approach to unify feature transformation suitable for both 3D-to-2D and 2D-to-3D VT, coupled with pre-computation for speed enhancement. Leveraging CNN-based probabilistic correspondences, HeightTrans and Prob-LSS extend the capabilities of respective VT methods effectively. Through one-stage fusion of dual features using DFF, DualBEV captures the essence of VT and demonstrates the effectiveness of unveiling precise correspondences for BEV representation. Furthermore, our approach is versatile and applicable to tasks such as BEV segmentation or 3D occupancy prediction.

However, our framework's current design exclusively derives all probabilities from the current frame, neglecting historical information and underutilizing the temporal module. Furthermore, our framework heavily depends on depth estimation, as demonstrated by a significant performance drop observed after switching to a uniform distribution, as shown in \cref{tab:Ablation Studt}(c).


\section*{Acknowledgements}
The authors would like to thank reviewers for their detailed comments and
instructive suggestions. This research is supported by National Key R\&D Program of China (2022YFB4300300).

%
%
\bibliographystyle{splncs04}
\bibliography{egbib}
\end{document}